\documentclass[11pt,a4paper]{article}
\usepackage[hyperref]{acl2021}
\usepackage{times}
\usepackage{latexsym}
\usepackage{booktabs}
\usepackage{adjustbox}
\usepackage{comment}

\newcommand{\af}[1]{\textcolor{blue}{#1}}

\usepackage{microtype}

\aclfinalcopy 

\title{Translate, \textit{then} Parse! 
A strong baseline for Cross-Lingual AMR Parsing}

\newcommand*\samethanks[1][\value{footnote}]{\footnotemark[#1]}
\author{Sarah Uhrig\thanks{*Equal contribution.} \space\space\space\space\space Yoalli Rezepka García\samethanks \space\space\space\space\space Juri Opitz \space\space\space\space Anette Frank \\
  Dept.\ of Computational Linguistics \\
  Heidelberg University \\
  69120 Heidelberg \\
 {\tt $\langle$surname$\rangle$@cl.uni-heidelberg.de} }

\date{}

\begin{document}
\maketitle
\begin{abstract}
In cross-lingual Abstract Meaning Re\-pre\-sen\-ta\-tion (AMR) parsing, researchers develop models that project sentences from various languages onto their AMRs to capture their essential semantic structures:
given a sentence in any language, we aim to capture its
core semantic content through concepts connected by manifold types of semantic relations.
Methods typically leverage large silver training data to learn a single
model that is able to project non-English sentences to AMRs.
However, we find that a simple baseline tends to be overlooked: translating the sentences to English and projecting their AMR with
a monolingual AMR parser
(\texttt{translate+parse}, \texttt{T+P}). 
In this
paper, we revisit
this simple two-step baseline, and enhance it
with a strong NMT system and a strong AMR parser. Our experiments show that 
\texttt{T+P} outperforms a recent state-of-the-art system across all tested languages: 
German, Italian, Spanish and Mandarin with +14.6, +12.6, +14.3 and +16.0 Smatch points.
\end{abstract}

\section{Introduction}

Abstract Meaning Representation (AMR), introduced by \citet{banarescu-etal-2013-abstract}, aims at representing the meaning of a sentence in a semantic graph format. Nodes represent entities, events and concepts, while (typed) edges express their relations. 

AMR itself, as of now, is 
English-focused, e.g., predicate frames are linked 
to
English PropBank \cite{kingsbury2002treebank}. 
However, the abstract nature of AMR, and  the fact that they are not explicitly linked to syntactic structure, make it appealing for extracting semantic structure of sentences in
various languages.
This insight led to the recent interest in a new task: \textbf{cross-lingual AMR parsing} \cite{damonte-cohen-2018-cross}. Here, researchers develop models to project sentences from different languages onto AMR graphs.
\begin{figure}
    \centering
    \includegraphics[width=0.8\linewidth]{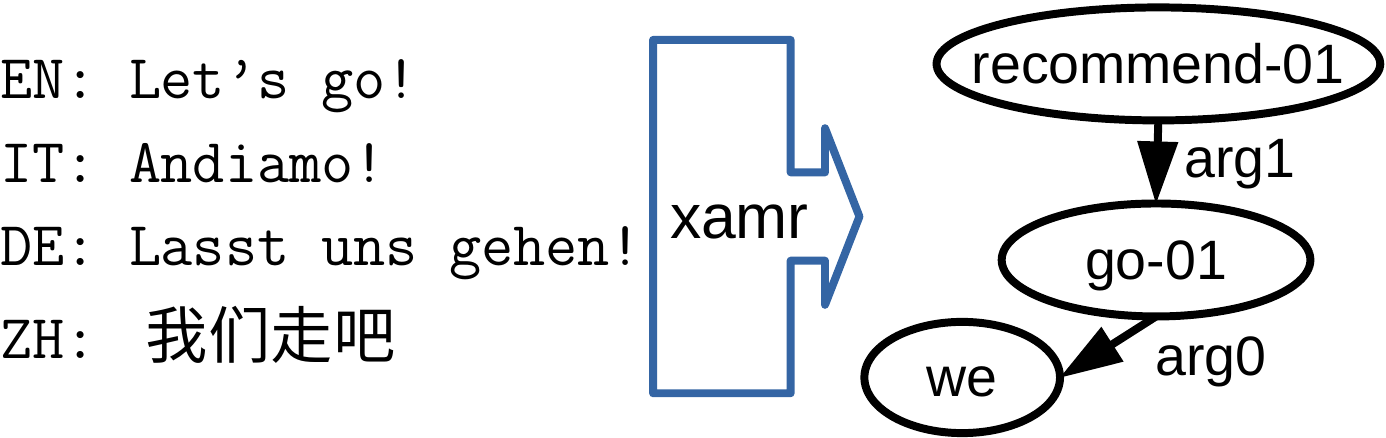}
    \caption{Cross-lingual AMR parsing as introduced by \citet{damonte-cohen-2018-cross}. 
    }
    \label{fig:ex1}
 
\end{figure}
Models that have recently been proposed are typically trained on large-scale silver data and learn to directly project the non-English sentences onto their AMR graphs (see Figure \ref{fig:ex1}) \cite{damonte-cohen-2018-cross, blloshmi-etal-2020-xl}. However, there is an intuitive baseline that we argue has so-far received too little attention: \texttt{translate+parse}, \texttt{T+P}. It first translates a sentence to a pivot language and applies
a mono-lingual parser for that language.
In light of the rapid 
progress of both NMT and 
AMR parsing models for English, our hypothesis is that this baseline has become more effective and  thus more realistic. Moreover, we argue that it could be beneficial to  
disentangle
two key
latent representations involved in the process of cross-lingual AMR parsing: i) 
one that translates between two natural languages and ii) one
that translates between a natural language and a 
meaning representation. This way, the cross-lingual AMR construction process is more transparent and can be better analyzed. 

In 
our work
we test these
hypotheses by translating the source language sentences into English with a strong NMT system, and 
parse 
the resulting English sentences using
a strong AMR parser. 
We show that our baseline delivers strong performance in cross-lingual AMR parsing across all considered languages, outperforming task-focused state-of-the-art models in all settings. We also discuss \textbf{fairer evaluation of cross-lingual AMR parsing} and \textbf{relevant implications} of this work for research into cross-lingual AMR parsing. 

We will release all code under public license.\footnote{\url{https://github.com/Heidelberg-NLP/simple-xamr}}

\section{Related work}

\paragraph{Cross-lingual AMR parsing} 

Cross-lingual AMR parsing was introduced by \citet{damonte-cohen-2018-cross}. They trained
an alignment-based AMR parser model that leverages large amounts of parallel silver AMR data obtained through
annotation projection from a curated parallel corpus. The authors also discussed
\texttt{translate+parse} (\texttt{T+P}) as a baseline using either the NMT systems Google translate and Nematus \cite{sennrich2017nematus}, or the SMT system Moses  \cite{koehn-etal-2007-moses}, together with a mono-lingual transition-based parser \cite{damonte-etal-2017-incremental}. However, their best 
\texttt{T+P} approach was Google Translate (GT) -- which cannot be fully replicated by  other researchers since both training data and model structure are hidden.
Given the recent advances in NMT \cite{barrault-etal-2019-findings, barrault-etal-2020-findings} and mono-lingual AMR parsing \cite{xu-etal-2020-improving}, where parsers now achieve scores on par with human IAA assessments  (c.f.\ \citet{banarescu-etal-2013-abstract}), we show that time is ripe
to put more spotlight on \texttt{T+P}.

\citet{blloshmi-etal-2020-xl} address the problem from complementary perspectives: i) they train a system that
projects AMR graphs from parsed English sentences to target sentences 
via
a parallel corpus, yielding \textit{gold non-English sentences} and \textit{silver AMRs}. Conversely, ii)
they train a system that
employs an NMT
system to translate English sentences from a human-annotated AMR dataset to another language, yielding pairs of \textit{silver non-English sentences} and \textit{gold AMRs}. This alleviates the dependency on external
AMR aligners. 

\paragraph{(Mono-lingual) AMR parsing} Mono-lingual AMR parsing equally made big strides in 
recent years, so that
today AMR parsers deliver benchmark scores that are
on-par with measured human IAA. The latest step forward was achieved with neural sequence-to-sequence models 
pre-trained on large-scale MT benchmark data \cite{roberts2020much, xu-etal-2020-improving} or 
are 
fine-tuning self-supervised seq-to-seq language models such as T5 or BART \cite{lewis2019bart, bevilacqua2021one}. Previou models perform parsing based on different techniques, e.g., predicting latent alignments jointly with nodes \citep{lyu-titov-2018-amr}, or via an iterative BFS writing traversal \citep{cai-lam-2019-core, cai-lam-2020-amr}. 

\section{Translate, then parse!  
}

Our pipeline model contains two components:

\paragraph{Sent-to-Sent: NMT system} \label{NMTSystem} We use Helsinki-NLP's \textit{Opus-MT models} \cite{tiedemann2020opus} to translate the sentences to
English
. The models are freely accessible\footnote{They are implemented in
\textit{EasyNMT}, a SOTA NMT package: \url{https://github.com/UKPLab/EasyNMT}} and provide high scores on public evaluation benchmarks.\footnote{See \url{https://huggingface.co/Helsinki-NLP} for scores on benchmarks.} 

\paragraph{Sent-to-AMR: AMR parser} For parsing English target sentences to AMR, we use the parser from \textit{amrlib}\footnote{\url{https://github.com/bjascob/amrlib}}, which consists of a T5 language model \cite{roberts2020much} that has been fine-tuned on English sentences and their AMRs. 

\section{Experiments}

\paragraph{Data} We employ the 
cross-lingual AMR parsing benchmark \textit{LDC2020T07}. 
It was built from the test 
split 
of the English mono-lingual LDC2017T10 data by translating its sentences to 
four languages: German, Spanish, Italian and Mandarin Chinese.
This amounts to
a total of 5,484 AMR-sentence pairs, or 
1,371 AMR-sentence pairs per language. 

\paragraph{Baselines} For all languages (German, Spanish, Italian and Mandarin Chinese), we compare against i) AMREAGER \cite{damonte-cohen-2018-cross}, and ii) XL-AMR \cite{blloshmi-etal-2020-xl}.

\paragraph{Evaluation metrics}
Our main evaluation metric is Smatch F1
\cite{cai-knight-2013-smatch}. The Smatch metric aligns the predicted graph with the gold graph and computes an F1 score that measures normalized triple overlap. Additionally, we calculate F1 scores for finer-grained core semantic sub-tasks \citet{damonte-etal-2017-incremental}.\footnote{i) Unlabeled: score without node labels, ii) No WSD: score w/o predicate sense disambiguation; iii) Reentrancies: score on re-entrant nodes (coreference); iv) Concepts: score on concept nodes; v) Named Ent.: indicating 
NER 
performance; vi) negation: polarity detection performance; vii) SRL: semantic role labeling performance.} In our analyses (\S \ref{par:gradedmetric}), we also study results with S2MATCH \cite{opitz-etal-2020-amr}, that offers a potentially fairer evaluation in cross-lingual AMR parsing, since it does not penalize allowed paraphrases that may emerge, e.g., due to the non-monotonous nature of translation (e.g., \textit{huckleberry} $\rightarrow$ \textit{Heidelbeere} (DE) $\rightarrow$ $blueberry$).

\begin{table*}[ht]
\centering
\scalebox{0.70}{
\begin{tabular}{l|llll|llll|llll}
             & \multicolumn{4}{c|}{AMREAGER} & \multicolumn{4}{c|}{XL-AMR} & \multicolumn{4}{c}{\texttt{translate+parse}} \\ \toprule
Metric       & DE    & ES    & IT    & ZH    & DE      & ES     & IT     & ZH     & DE    & ES    & IT    & ZH    \\ \midrule
SMATCH       & 39.1  & 42.1  & 43.2  & 34.6  & 53.0    & 58.0   & 58.1   & 43.1   & \textbf{67.6} & \textbf{72.3}  & \textbf{70.7}    & \textbf{59.1}     \\ \midrule
Unlabeled    & 45.0  & 46.6  & 48.5  & 41.1  & 57.7    & 63.0   & 63.4   & 48.9   & \textbf{71.9} & \textbf{76.5}  & \textbf{75.1}     & \textbf{65.4}     \\
No WSD       & 39.2  & 42.2  & 42.5  & 34.7  & 53.2    & 58.4   & 58.4   & 43.2   & \textbf{67.9} & \textbf{72.7}  & \textbf{71.1}     & \textbf{60.4}     \\
Reentrancies & 18.6  & 27.2  & 25.7  & 15.9  & 39.9    & 46.6   & 46.1   & 34.7   & \textbf{55.8} & \textbf{60.9}   & \textbf{58.2}     & \textbf{47.5}     \\
Concepts     & 44.9  & 53.3  & 52.3  & 39.9  & 58.0    & 65.9   & 64.7   & 48.0   & \textbf{71.4} & \textbf{78.1}  & \textbf{75.6}     & \textbf{63.3}     \\
Named Ent.   & 63.1  & 65.7  & 67.7  & 67.9  & 66.0    & 66.2   & 70.0   & 60.6   & \underline{\textbf{86.3}} & \underline{\textbf{86.6}}   & \textbf{87.6}     & \underline{\textbf{84.2}}     \\
Negation     & 18.6  & 19.8  & 22.3  & 6.8   & 11.7    & 23.4   & 29.2   & 12.8   & \underline{\textbf{49.0}} & \underline{\textbf{59.5}}   & \underline{\textbf{55.7}}     & \textbf{38.5}     \\
SRL          & 29.4  & 35.9  & 34.3  & 27.2  & 47.9    & 55.2   & 54.7   & 41.3   & \textbf{61.7} & \textbf{68.0}   & \textbf{65.8}     & \textbf{54.1}   \\
\bottomrule
\end{tabular}
}
\caption{F1 Smatch for two baselines and \texttt{T+P}. Best results in \textbf{bold}. Improvements $>$ 20 points are \underline{underlined}.}
\label{tab:results}
\end{table*}

\subsection{Main results} Results are displayed in Table \ref{tab:results}. Overall, our \texttt{translate+parse} baseline
outperforms previous work by large margins. In all assessed semantic categories, 
\texttt{T+P}
outperforms XL-AMR models by more than 10 Smatch points.  The smallest improvement obtained is achieved in IT with +12.6 points.

In some key semantic categories, the differences are extreme. For negation detection we obtain performance improvements that range from +26.5 points (IT) to +37.3 (DE). The named entity recognition improves by +20.3 points for German, +20.4 points for Spanish, +17.6 points for Italian.

\begin{table*}[t]
    \centering
    \scalebox{0.68}{    \begin{tabular}{ll|llll|llll}
    && \multicolumn{4}{c}{XL-AMR} & \multicolumn{4}{c}{\texttt{translate+parse}}\\
     \toprule
    \multicolumn{2}{c}{Metric} & DE    & ES    & IT    & ZH & DE    & ES    & IT    & ZH\\
    \midrule
    $\tau$=$\frac{1}{2}$&S$2$M P &59.7(4.3)&63.9(+3.9)&64.7(\textbf{+4.7})&49.2(\textbf{+4.7})& 74.1(+3.1) & 78.3(+2.5) &  77.0(+2.8) & 65.8 (\textbf{+4.0})  \\
    & S$2$M R &54.8(+4.0)&59.9(+3.7)&59.4(+3.7)&47.3(\textbf{+5.0})& 67.3(+2.9) & 71.5(+2.4) & 70.0(+2.5) & 60.4 (\textbf{+3.7})\\
    \cmidrule{2-10}
    & S$2$M F1 &57.1 (+4.1)&61.8(+3.1)&62.0(+3.9)&48.2(\textbf{+5.1})& 70.5(+2.9) & 74.7(+2.4) & 73.4(+2.7) & 63.0 (\textbf{+3.9}) \\

    \midrule
    \midrule
    
    $\tau$=0.0&S$2$M P &61.5(+6.1)&65.4(+5.4)&66.2(+6.2)&51.3(\textbf{+6.6})& 75.2(+4.2) & 79.1(+3.3) &  77.9(+3.7) & 67.1 (\textbf{+5.3})  \\
    & S$2$M R &56.4(+5.6)&61.2(+5.0)&60.8(+5.1)&49.4(\textbf{+7.1})& 68.2(+3.8) & 72.1(+3.0) & 70.8(+3.3) & 61.6 (\textbf{+4.9})\\
    \cmidrule{2-10}
    & S$2$M F1 &58.9(+5.9)&63.2(+5.2)&63.4(+5.3)&50.4(\textbf{+7.3})& 71.5(+3.9) & 75.4(+3.1) & 74.2(+3.5) & 64.3 (\textbf{+5.2}) \\
    
    \bottomrule
    \end{tabular}}
    \caption{Evaluation using a graded metric. \textbf{Bold}: Largest improvement of a parser using fairer graded evaluation.}
    \label{tab:s2matchres}
\end{table*}

\begin{table}
    \centering
    \scalebox{0.5}{
    \begin{tabular}{lrrrrrr|rrrrrr}
    & \multicolumn{6}{c}{XL-AMR} & \multicolumn{6}{c}{\texttt{translate+parse}}\\
    & D-ES & D-I & D-Z & ES-I & ES-Z & I-Z & D-ES & D-I & D-Z & ES-I & ES-Z & I-Z \\
    \toprule
    SMATCH & 52.3 & 52.2 & \textit{40.9} & \textbf{58.5} & 42.7 & 42.8 & 74.3 & 73.7 & \textit{61.6} & \textbf{79.0} & 63.7 & 63.1  \\
    S2MATCH & 58.7 & 58.6 & \textit{48.7} & \textbf{63.9} & 50.2 & 50.4 & 77.7 & 77.2 & \textit{66.7} & \textbf{81.8} & 68.6 & 68.0  \\
    \midrule
Unl.\ & 57.0 & 57.2 & \textit{46.6} & \textbf{63.3} & 48.0 & 48.7 & 77.0 & 76.7 & \textit{65.7} & \textbf{81.6} & 67.8 & 67.1  \\
NoWSD & 52.3 & 52.4 & \textit{41.0} & \textbf{58.7} & 42.7 & 42.8 & 74.4 & 73.8 & \textit{61.7} & \textbf{79.0} & 63.8 & 63.1  \\
Conc.\ & 57.7 & 56.8 & \textit{45.0} & \textbf{65.4} & 47.7 & 47.4 & 75.1 & 74.2 & \textit{63.1} & \textbf{80.0} & 65.6 & 64.8  \\
NER & 63.9 & 64.3 & \textit{56.6} & \textbf{68.2} & \textit{56.6} & 57.9 & 90.1 & 90.1 & \textit{83.8} & \textbf{91.2} & 84.0 & 84.2  \\
Neg.\ & 7.6 & 9.7 & \textit{6.6} & \textbf{44.0} & 12.8 & 11.6 & 61.3 & 55.4 & \textit{42.}3 & \textbf{69.4} & 46.2 & 47.8  \\
Reent.\ & 40.7 & 41.5 & \textit{34.6} & \textbf{49.8} & 36.7 & 37.2 & 64.2 & 63.2 & \textit{50.4} & \textbf{69.6} & 51.8 & 51.0  \\
SRL & 47.4 & 47.5 & \textit{39.5} & \textbf{55.8} & 41.7 & 41.7 & 69.7 & 68.9 & \textit{56.5} & \textbf{75.0} & 58.3 & 57.6  \\
\bottomrule
    \end{tabular}}
    \caption{Semantic consistency over language pairs measured with SMATCH, S2MATCH ($\tau$=$0$). \textbf{Bold}/\textit{italics
    }: highest/lowest score for language pairs.}
    \label{tab:consistent}
\end{table}

\subsection{Studies}

\paragraph{Using a graded metric for evaluation}\label{par:gradedmetric} When evaluating predicted AMRs against reference AMRs in cross-lingual AMR parsing, we are essentially 
comparing AMRs from sentences that are not exactly the same. This means that predicted concepts that are valid may 
get erroneously penalized by the evaluation metric. For instance, consider a
German source sentence that contains \textit{Heidelbeere}, and our cross-lingual AMR system predicts i) \textit{huckleberry} or ii) \textit{blueberry}. Depending on which concept is mentioned in the reference AMR graph (based on
the unseen sentence from which the human SemBank annotator created this graph), only one of the two options will be viewed as correct, which results in unfair evaluation. To mitigate this, we 
propose to conduct the cross-lingual AMR evaluation using S$2$MATCH  \cite{opitz-etal-2020-amr}, a metric that admits
graded concept similarity.
S$2$MATCH has a hyper-parameter $\tau$ that sets a threshold for sufficiently similar concept nodes across AMRs, using cosine-similarity. The alignment of similar concepts can increase the final score.
The default $\tau$ is 0.5, but we also try 0.0 which is less strict and fosters dense alignment.

\begin{figure}
    \centering
    \includegraphics[width=0.7\linewidth]{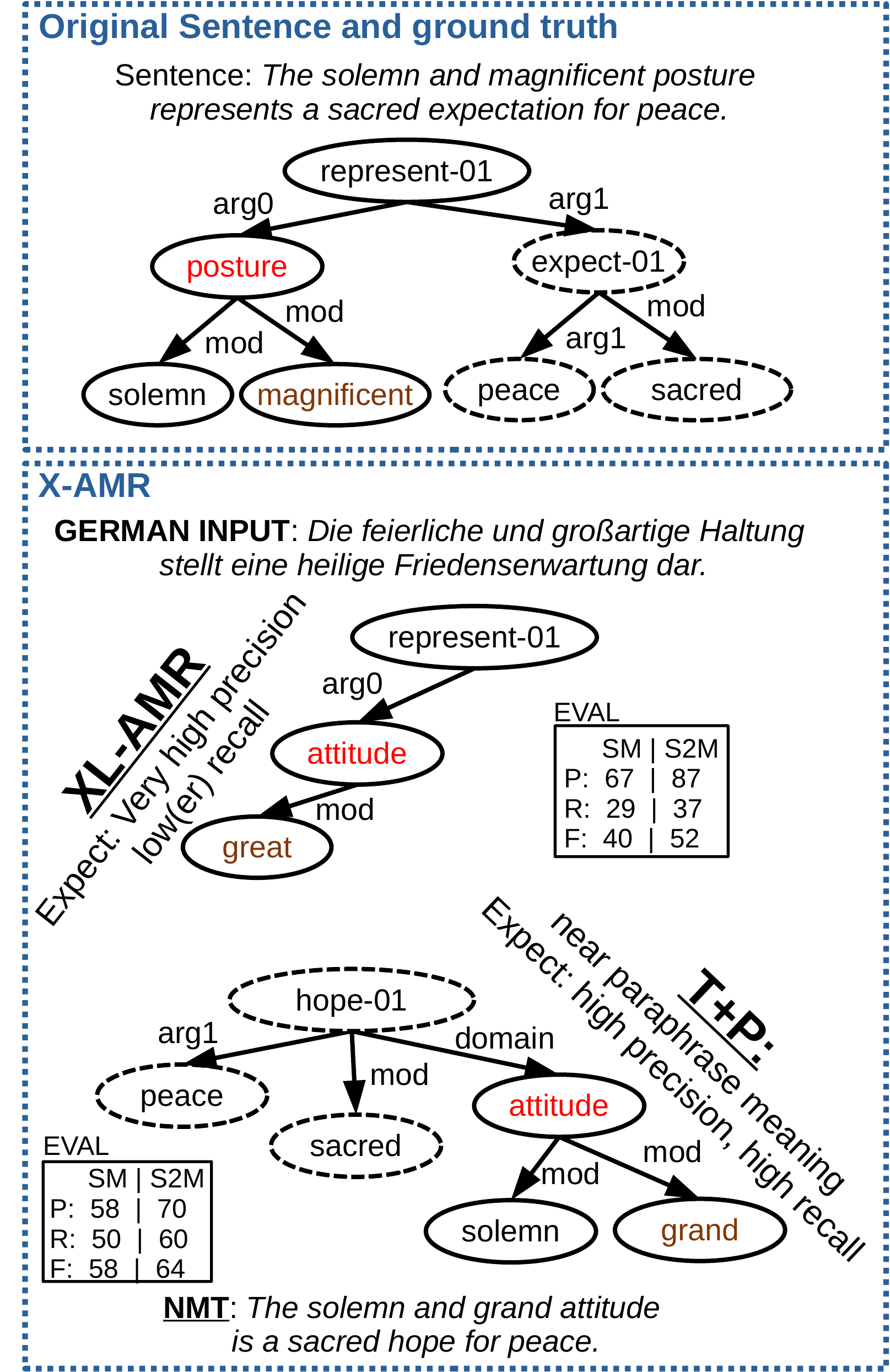}
    \caption{Evaluation example.}
    \label{fig:ex2}
\end{figure}

The results are displayed in Table \ref{tab:s2matchres}. Interestingly, most score improvements are obtained for German (+3.9 points) and Mandarin Chinese (+5.2 points). We conjecture that this is because there is slightly less variety in EN-$\{$ES, IT$\}$ translations, than for EN-DE, and especially for EN-ZH. This is also visible from the results of our baseline XL-AMR, which we reevaluate using S2MATCH: Most gains are obtained for Mandarin Chinese with an improvement of more than 7 points F1 score. Inspecting test cases manually, we find many cases were S2MATCH made the evaluation fairer. For instance, the following \textit{gold-pred (DE: [German word])} concept tuples are ignored by SMATCH but considered by S2MATCH: \textit{pledge-promise (DE: `versprechen');  write-compose (DE: `verfasst') strong-resolute (DE: `deutlich'); spirit-ghost (DE: `Geist'),} etc. In all these cases the cross-lingual AMR system predicted the correct concept, but was penalized by SMATCH. A concrete example case, with lexical (see colored nodes) and structural (see dotted nodes) meaning-preserving divergences, is shown in Fig.\ \ref{fig:ex2}. 

For future work that applies cross-lingual AMR parsing evaluation, we recommend additional evaluation assessment with S2MATCH.

\paragraph{NMT quality} The quality of our automatic translations is evaluated with two metrics: i) BLEU score \cite{papineni-etal-2002-bleu} and ii) S(entence-)BERT \cite{reimers-gurevych-2019-sentence}, in order to assess surface-oriented
as well as 
semantic similarity. For SBERT, we create sentence embeddings for both our translations and the English reference sentences and compute pair-wise cosine 
similarity.

\begin{table}[]
    \centering
    \scalebox{0.7}{
    \begin{tabular}{l|llll|l}
    \toprule
         & DE & ES & IT & ZH & mean \\
         \midrule
      BLEU & 0.41 & 0.49 & 0.46 & 0.23 & 0.40\\
      SBERT (cosim)   & 0.93 & 0.95 & 0.94 & 0.88 & 0.92\\
      \bottomrule
    \end{tabular}}
    \caption{BLEU and SBERT MT quality assessment.}
    \label{tab:translationresults}
\end{table}

Looking at the quality of our MT outputs
(Table \ref{tab:translationresults}), we see that 
translation quality is generally quite high. The moderate BLEU scores seem to result 
more from variation in surface form
than from incorrect translations, which is backed by the high cosine similarity scores across languages (and also highlights the need for a fairer and graded AMR evaluation as 
proposed 
above.\footnote{This is also supported by a manual analysis of samples.}  Finally, comparing the different source languages, there seems to be a higher quality in the translations from German, Spanish, and Italian, 
compared to Mandarin Chinese. This is not only reflected in the BLEU scores, but also in the SBERT cosine scores, which suggest a higher semantic similarity between our translations from DE, ES, IT and the reference sentences. 

\paragraph{Semantic cross-lingual consistency of cross-lingual AMR systems} A cross-lingual AMR system should be expected to deliver the same or highly similar AMRs for two sentences from different languages, if the sentences carry the same meaning. We may say that a system is \textit{semantically consistent} if it complies to this expectation. 

To measure the degree of consistency, we evaluate 
system outputs of a cross-lingual AMR system for input language X against the outputs of the same system when fed sentences in
language Y, 
from a \textit{parallel} dataset (X,Y) of sentences in 
languages X and Y. In the standard evaluation, we computed \textit{EVAL}$(system(X), A)$ and \textit{EVAL}$(system(Y), A)$, where $A$ are target AMRs. In this experiment, we instead calculate \textit{EVAL}$(system(X), system(Y))$, assessing the degree of consistency of a $system$.

The results are provided in Table \ref{tab:consistent}, where we see a very clear picture that holds true both for our joint baseline (XL-AMR) and our \texttt{T+P} approach and all examined semantic categories: the highest consistency is achieved for Spanish-Italian (ES-I, XL-AMR: 63.9 S2MATCH; \texttt{T+P}: 81.8 S2MATCH), while the lowest consistency is achieved for German and Mandarin Chinese (D-Z, XL-AMR: 48.7 S2MATCH; \texttt{T+P}: 66.7 S2MATCH). When directly comparing the parsing systems, overall \texttt{T+P} appears to offer better consistency in all categories, especially negation. However, the substantial variance between languages may indicate that either i) there is a great necessity for making cross-lingual parsers more robust or, ii), 
that 
AMR representations, as constructed from English, may be better prepared to represent (besides English) Spanish and Italian language, than, e.g, German or Chinese.

\section{Discussion}

We believe that the surprising effectiveness of \texttt{translate+parse} touches upon a key question: \textbf{to what degree can AMR be considered 
an interlingua}? On one hand, \citet{banarescu-etal-2013-abstract} explicitly state that AMR `is not designed as an interlingua'. 
Indeed, AMRs created for English sentences do have a
flavour of English, since they are
partially grounded in English PropBank \cite{kingsbury2002treebank}. But linking AMRs to 
a PropBank of another language, e.g., Brazilian \cite{duran2012propbank} or Arabic \cite{palmer2008pilot}, and parsing non-English sentences into corresponding AMRs, would not solve, but only displace the problem of being tied to a specific language's lexical semantic inventory.\footnote{Potentially, this may be mitigated in the future by linking AMR to x-lingual PropBanks \cite{akbik-etal-2015-generating}} On the other hand, 
AMR \textit{does} contain 
abstract meaning components that represent
language phenomena 
we may consider as universals: negation, occurrence of named entities, semantic events and their related participants, as well as semantic relations such as Possession, Purpose or Instrument.\footnote{C.f.\ \citet{xue-etal-2014-interlingua}.} We argue that this abstract structure again pushes AMR more towards an interlingua. Hence, the emergent interest in cross-lingual (A)MR \cite{ oepen-etal-2020-mrp, fan-gardent-2020-multilingual, sheth-etal-2021-bootstrapping, sherborne2021zeroshot} is well justified. However, even if AMR's \textit{inventory}  may favor an interlingual representation, we cannot, in general, expect a homomorphism of AMRs constructed from semantically equivalent sentences in various languages,
given wide-spread phenomena that can preclude a uniform AMR representation, such as constructions
involving 
head-switching phenomena or differences in lexical meaning. 


Such a middle-ground is 
indicated by our results:
(Too) much divergence may be involved when mapping non-English sentences to original \texttt{EN}-AMRs \textit{directly}, which is penalized by the strict(er) SMATCH metric. We show that evaluation with the softer S2MATCH metric admits small deviations in the conceptual inventory of different languages.
The fact that our
\textit{indirect} two-step approach \texttt{T+P} shows very strong performance also strengthens the view that AMR is not fully an interlingua. The better performance of \texttt{T+P}
may in part
be due to a capacity of strong NMT systems to neutralize some amount of inter-lingual divergence, so that evaluation against \texttt{EN}-AMRs can yield better results in this setting.

Note that in our  \texttt{T+P} approach
two important intermediate (latent) representations are clearly separated:
one in the NMT model (that builds a bridge between two natural languages) and 
one in the parser (that builds a bridge between English
and a language of meaning with a flavor of English). 
By analyzing divergences between source and target in the \texttt{T} step, we can uncover aspects of semantic representations
that are \textit{not} isomorphic between languages, and which -- by transfer via translation -- may be neutralized to match the pivot-flavored AMR structure. 
Hence, the \texttt{T+P} approach offers an ideal framework for studying interlingual similarities and divergences in cross-lingual AMR parsing, by comparing the structural-semantic divergences of non-English sentences and their translated English counterparts (aka translational divergences), with the aim of identifying structural-semantic differences between languages that can affect the cross-lingual mapping of sentences into a uniform interlingual AMR.\footnote{For example, we may train a system to parse Non-EN sentences to EN-(flavored)AMR graphs, and compare them to AMRs we obtain from \textit{translated EN sentences} targeting the same EN-AMRs. Divergences we find between AMRs predicted in these settings can indicate phenomena leading to non-isomorphic AMRs that require attention when aiming for a true interlingual AMR formalism.}

\section{Conclusion}

We revisited \texttt{translate+parse}, an intuitive baseline for cross-lingual AMR parsing. 
Equipped with a recent NMT system and a monolingual AMR parser,
\texttt{T+P} outperforms other approaches by large margins across all evaluation settings. We propose to 
employ a 
graded metric for fairer evaluation of cross-lingual AMR parsing. Our work 
can
serve as a strong base\-line for future development of 
cross-lingual AMR parsers. 
Finally, the \texttt{T+P} approach provides an ideal platform for 
deeper assessment, analysis, and break-down of potential interlingual aspects of AMR.

\section*{Acknowledgments}

This  work  has  been  partially funded  by  the  DFG  within the project ACCEPT as part of the Priority Program  “Robust  Argumentation  Machines”  (SPP-1999).  
We thank our anonymous reviewers for constructive comments and valuable feedback.


\bibliographystyle{acl_natbib}
\bibliography{anthology,acl2021}


\end{document}